\documentclass{article}
\usepackage{tikz}
\usepackage{natbib}

\usepackage[utf8]{inputenc}  
\usepackage{fontenc}
\usepackage{hyperref}
\usepackage{amsmath}
\usepackage{amsfonts}
\usepackage{lscape}
\usepackage{float}
\usepackage{a4wide}
\usepackage{dsfont}
\usepackage{bm}
\usepackage{comment}

\usetikzlibrary{shapes,arrows}

\usepackage{bm}
\usepackage{color}
\usepackage{amssymb}

\usepackage{amsmath}
\usepackage{url}
\usepackage{float}
\usepackage{cancel}

\usepackage{multirow}

\usepackage{mathtools}
\usepackage{witharrows}
\usepackage[ruled,vlined]{algorithm2e}

\SetCommentSty{mycommfont}

\usepackage{xargs}     
\usepackage[colorinlistoftodos,prependcaption,textsize=tiny]{todonotes}
\newcommandx{\improvement}[2][1=]{\todo[linecolor=Plum,backgroundcolor=Plum!25,bordercolor=Plum,#1]{#2}}
\usepackage{booktabs}
\usepackage{arydshln}

\usepackage{natbib}
\newcommand{\mb}[1]{\mathbf{#1}}

\newcommand{\R}{\mathbb R}
\newcommand{\btheta}{\pmb{\theta}}

\newcommand{\Expectation}[2]{\mathds{E}_{#1}\left[#2\right]}
\newcommand{\KL}[2]{\text{\KLD} \left( {#1} \parallel {#2}\right)}

\newcommand{\complexity}[1]{\mathcal{O}(#1)}

\newcommand{\Ngauss}[1]{\mathcal{N}\left(#1\right)}

\newcommand{\Xspace}{\mathcal{X}\xspace} 
\newcommand{\X}[1]{\mb{x}_{#1}\xspace} 
\newcommand{\Xsamples}{\mb{X}\xspace} 

\newcommand{\Z}[1]{\mb{z}_{#1}\xspace} 
\newcommand{\Zsamples}[1]{\mb{Z}^{#1}\xspace}

\newcommand{\Ysamples}{\mb{y}\xspace} 

\newcommand{\varm}[1]{\mathbf{m}^{#1}}
\newcommand{\varS}[1]{\mathbf{S}^{#1}}

\newcommand{\myprod}[2]{\prod_{#1}^{#2}} 
\newcommand{\mysum}[2]{\sum_{#1}^{#2}}

\newcommand{\dd}[1]{\text{d}#1}

\newcommand{\W}[1]{\mb{W}^{#1}\xspace}

\newcommand{\f}[1]{\mb{f}_{#1}\xspace} 

\newcommand{\fK}[1]{\mb{f}^{#1}_K\xspace} 
\newcommand{\fnull}[1]{\mb{f}^{#1}_0\xspace} 
\newcommand{\fpos}[2]{\mb{f}^{#2}_{#1}\xspace}

\newcommand{\uu}[1]{\mb{u}^{#1}\xspace} 

\newcommand{\uK}[1]{\mb{u}^{#1}_K\xspace} 
\newcommand{\unull}[1]{\mb{u}^{#1}_0\xspace}

\newcommand{\idparams}[2]{\btheta(\W{#1},#2)^{#1}}
\newcommand{\Wspace}{\mathcal{W}}

\newcommand{\G}{\mathbb{G}}
\newcommand{\Gk}[1]{\mathbb{G}_{#1}}
\newcommand{\Gcomp}{\Gk{\btheta}}

\newcommand{\Gc}[1]{\mathbb{G}^{#1}}

\newcommand{\Gcompc}[1]{\Gk{\btheta}^{#1}}

\newcommand{\Gcompcinpdep}[2]{\Gk{\btheta(\W{#1},#2)}^{#1}}

\newcommand{\T}{\mathbb{T}}

\newcommand{\Jacobian}[1]{\myprod{k=0}{K-1}\left| \det \frac{\partial \G_{\theta_k}(#1)}{\partial #1} \right|^{-1}}

\newcommand{\Jacobiancinputdep}[3]{\myprod{k=0}{K^{#2}-1}\left| \det \frac{\partial \Gc{#2}_{\theta_k(\W{#2},#3)}(#1)}{\partial #1} \right|^{-1}}

\newcommand{\J}[1]{\pmb{J}_{#1}}

\newcommand\deeptransformedset{\{\fK{l}, \uK{l}\}_{l=1}^L}

\newcommand\setWs{\{\W{l}\}_{l=1}^L}
\definecolor{darkpink}{rgb}{0.91, 0.33, 0.5}
\definecolor{darkgreen}{rgb}{0.0, 0.5, 0.0}

\newcommand{\fig}{Fig.\xspace}

\newcommand{\acro}[1]{\textsc{#1}\xspace}

\newcommand{\GP}{\acro{\smaller GP}}
\newcommand{\ETGP}{\acro{\smaller ETGP}}
\newcommand{\TGP}{\acro{\smaller TGP}}
\newcommand{\DGP}{\acro{\smaller DGP}}
\newcommand{\IP}{\acro{\smaller IP}}
\newcommand{\DSVI}{\acro{\smaller DSVI}}
\newcommand{\ID}{\acro{\smaller ID}}

\newcommand{\DTGP}{\acro{\smaller DTGP}}
\newcommand{\NN}{\acro{\smaller NN}}

\newcommand{\UCI}{\acro{\smaller UCI}}

\newcommand{\ELBO}{\acro{\smaller ELBO}}
\newcommand{\KLD}{\acro{\smaller KL}}
\newcommand{\ELL}{\acro{\smaller ELL}}
\newcommand{\SVGP}{\acro{\smaller SVGP}}

\newcommand{\MCMC}{\acro{\smaller MCMC}}

\newcommand{\NLL}{\acro{\smaller NLL}}

\newcommand{\LOTUS}{\acro{\smaller LOTUS}}

\newcommand{\CPU}{\acro{\smaller CPU}}

\newcommand{\aka}{a.k.a.\xspace}

\newcommand{\eg}{e.g.\xspace}
\newcommand{\ie}{i.e.\xspace}

\newtheorem{definition}{Definition}

\title{Deep Transformed Gaussian Processes}

\author{
	Francisco Javier Sáez-Maldonado\footnote{Equal contribution.}\\
Universidad de Granada\\
\texttt{fjaviersaezm@ugr.es} \\
\and 
Juan Maro\~nas$^*$\\
Universidad Aut\'onoma de Madrid\\
\texttt{juan.maronnas@uam.es}  \\
\and 
Daniel Hern\'andez-Lobato\\
Universidad Aut\'onoma de Madrid\\
\texttt{daniel.hernandez@uam.es} 
}

\date{}

\begin{document}
	
\maketitle

\begin{abstract} 
Transformed Gaussian Processes (\TGP{}s) are stochastic processes specified by transforming samples from the joint distribution from a prior process (typically a \GP{}) using an invertible transformation; increasing the flexibility of the base process.
 Furthermore, they achieve competitive results compared with Deep Gaussian Processes (\DGP{}s), which are another generalization constructed by a hierarchical concatenation of \GP{}s. In this work, we propose a generalization of \TGP{}s named Deep Transformed Gaussian Processes (\DTGP{}s), which follows the trend of concatenating layers of stochastic processes. More precisely, we obtain a multi-layer model in which each layer is a \TGP. This generalization implies an increment of flexibility with respect to both \TGP{}s and \DGP{}s. Exact inference in such a model is intractable. However, we show that one can use variational inference to approximate the required computations yielding a straightforward extension of the popular \DSVI inference algorithm \citep{Salimbeni2017}. The experiments conducted evaluate the proposed novel \DTGP{}s in multiple regression datasets, achieving good scalability and performance.
\end{abstract}

\section{Introduction}
\label{section1:INTRO}

Although neural networks present highly accurate results on classification and regression tasks, they do not offer uncertainty estimations associated with the predictions made, which are mandatory in some fields. For example, in medicine, a typical problem is cancer detection using histopathological images, where given an image (or a set of images), the models try to determine if the patient has cancerous tissue or not \citep{schmidt2023probabilistic}. In this case, the model must offer an interpretable (\ie well-calibrated) output for the pathologists, who may use the output of the model for diagnosis. Bayesian learning offers a solution for this problem as it automatically outputs an estimation of the associated prediction uncertainty. However, it comes at a cost: computational tractability. The posterior distribution is most of the time intractable, so it has to be approximated using techniques such as variational inference. 

\emph{Gaussian Processes} (\GP{}s) are a very powerful, non-parametric model that allows inference in the function space by placing a prior distribution on the target latent function \citep{gps_for_ml_rasmussen}. These models have been studied extensively in the literature \citep{pmlr-v5-titsias09a, GPs-Big-Data}, leading to different generalizations. The most popular are \emph{Deep Gaussian Processes} (\DGP{}s) \citep{pmlr-v31-damianou13a}, which use the output of a \GP{} as the input to another \GP{}, increasing the expressiveness of the resulting model. Also, the usage of transformations on the prior and likelihood of the \GP{} has been explored. Using \emph{Normalizing Flows} \citep{rezende2016variational}, the \emph{Transformed Gaussian Process} (\TGP{}) \citep{maronas2021transforming} extends standard \GP{}s by transforming  the prior distribution of the \GP{}, which is no longer Gaussian, using an invertible transformation. 

In this work, \emph{Deep Transformed Gaussian Processes} (\DTGP{}s) are introduced as a novel form of extending Transformed Gaussian Processes by concatenating the output of a \TGP to another \TGP. This model aims to improve the performance and the uncertainty estimation of previous models by being more flexible than them and thus being able to model more complex data. Inference in \DTGP{}s is intractable, so the Evidence Lower Bound  is approximated using Monte Carlo samples, using a straightforward extension of the algorithm  proposed for inference in \DGP{}s by \cite{Salimbeni2017}. The usage of normalizing flows between each layer, leads to an increment in flexibility, adding a small term to the total computational cost.

To validate the proposed model, we conduct extensive experimentation with \DTGP{}s. With this goal, we employ a specifically designed toy dataset and eight real datasets from the \UCI repository, which have been used to test the performance of \DTGP{}s in comparison with other state-of-the-art models. Our results show that \DTGP{}s obtain better or comparable results in almost all datasets.

\section{Preliminaries}
\label{section2:PRELIM}

The problem we aim to solve is to infer an unknown function \(f: \R^D \to \R\) given noisy observations \(\Ysamples = (y_1\cdots,y_N)^T\) at locations \(\Xsamples = (\X{1},\cdots, \X{N})\), where in our problem $\X{} \in\Xspace \subseteq \mathbb{R}^D$. Gaussian Processes place a prior on \(f\) such that the distribution of all function values is jointly Gaussian, with a mean function \(\mu: \mathcal X \to \R\) and covariance function \(K: \mathcal X \times \mathcal X \to \R\) \citep{gps_for_ml_rasmussen}. Since the computational cost of exact inference in \GP{}s scales cubically with \(N\), a set of inducing locations \(\Zsamples{} = (\Z{1},\cdots,\Z{M})\) with \(M << N\) is considered \citep{pmlr-v5-titsias09a}, aiming to reduce this cost. We denote \(\f{} = f(\Xsamples)\) and \(\uu{} = f(\Zsamples{})\). The joint distribution of \(\f{}, \uu{}\) and \(\Ysamples\) is
\begin{equation}
    p(\Ysamples, \f{}, \uu{}) = \underbrace{\prod_{i=1}^N p(y_i \mid f_i)}_{\text{likelihood}} \underbrace{p(\f{} \mid \uu{}; \ \Xsamples, \Zsamples{}) p( \uu{} ; \ \Zsamples{})}_{\text{\GP prior}},
\end{equation}
where \(\uu{}\) is supposed to follow the same \GP prior as \(\f{}\): \(p(\uu{}) = \mathcal N (\uu{} \mid m(\Zsamples{}), K(\Zsamples{}, \Zsamples{}))\).\\

Some authors have explored the possibility of transforming the prior or the likelihood distributions \citep{maronas2021transforming}. 
In this sense, using two mappings \(\T,\G\), the whole modelling process can be generalized as
\begin{equation}
\begin{rcases}        \fnull{} \sim \mathcal{GP} (\mu(\Xsamples), K(\Xsamples, \Xsamples));  &\ \fK{} = \G(\fnull{}) \\        \T(\Ysamples) = \fK{} + \epsilon; &\epsilon \sim \mathcal N(0, \Sigma)
\end{rcases}
\end{equation}

\noindent In this work, we are interested in transforming the prior using \(\G\) and maintaining \(\T\) as the identity. For this we consider  $\G$ to be given by an invertible transformation (\aka Normalizing Flow), as in \cite{maronas2021transforming}, followed by a hierarchical concatenation  as in \cite{pmlr-v31-damianou13a}. However, from now on $\G$  will just denote the invertible transformation. Since the composition of invertible transformation remains invertible and differentiable, we define $\G$ to be the composition of $K$ invertible transformations \(\Gcomp = \Gk{0} \circ \Gk{1} \circ \cdots \circ \Gk{K-1}\). This composition helps to increase the flexibility of the flow as much as we want. Also, in each of the steps of the flow the parameters \(\pmb{\theta}_k\) may depend on the input via a transformation such as a Neural Network (\NN), giving rise to Input-Dependent (\ID) normalizing flows, which yield a non-stationary process with well-inductive biases \citep{maronas2022efficient,maronas2021transforming}. In this case, parameters are given by functions $\theta:\Wspace\times\Xspace \to \mathbb{R}$ which we denote as \(\idparams{}{\Xsamples}\).\\

The Transformed Gaussian Process (\TGP) is defined by the generative process given by composing a sample \(\f{0}\) of a \GP with a normalizing flow:
\begin{equation}
    \label{eq:tgp:definition}
    \fnull{} \sim \mathcal{GP}(\mu( \Xsamples), K( \Xsamples, \Xsamples)), \quad \fK{} = \Gcomp(\fnull{}).
\end{equation}
Following \citep{maronas2021transforming} we consider element-wise mappings, which produce diagonal Jacobians. This, together with the application of the inverse function theorem and change of variable formula, leads to the \TGP distribution: 
\begin{equation}\label{eq:tgp:transformed:distr}
p(\fK{} \mid \Gcomp , \Xsamples) = p(\fnull{} \mid \Xsamples) \Jacobian{\f{k}}.
\end{equation}
We denote \(\J{\pmb{a}}= \Jacobian{\pmb{a}}\). Unlike in standard \GP{}s, inference in \TGP{}s is intractable. The posterior distribution is efficiently approximated using a variational distribution \(q(\fK{},\uK{}) = p(\fK{} \mid \uK{}) q(\uK{})\), which is chosen to contain a factor equal to the exact conditional prior $p(\fK{} \mid \uK{})$ and a marginal variational distribution \(q(\uK{}) = \mathcal N(\unull{} \mid \varm{}, \varS{})\J{\uK{}}\). A cancellation of both the conditional prior and the Jacobian \(\J{\uK{}}\) is achieved in the variational \ELBO giving: 
\begin{equation}\label{eq:tgp:elbo:simple}
        \mathcal L_{\text{\TGP}} =  \Expectation{q(\fnull{})}{\log p(\Ysamples \mid \Gcomp (\fnull{})} -  \KL{q(\unull{})}{p(\unull{})}.
\end{equation}
This provides a bound that can be evaluated in \(\complexity{NM^2 + M^3 + NLK}\), where \(L\) is the computational cost of the normalizing flow. When \ID flows are used, a Bayesian treatment can be considered by assigning a prior distribution \(p(\W{})\) to the weights of the \NN, see \cite{maronas2021transforming}.

\section{Deep Transformed Gaussian Processes}
\label{section3:DTGP}

In this section, we present Deep Transformed Gaussian Processes (\DTGP{}s) which generalize \TGP{}s through their hierarchical composition. Following the fashion of \DGP{}s, the output of the \TGP is used as the input of another \TGP, recursively defining a stochastic process:
\begin{definition}[Deep Transformed Gaussian Process]
    A \emph{Deep Transformed Gaussian Process (\DTGP)} is a collection of random variables \(\{\fpos{K}{h,l}\}_{h=1,l=1}^{H^l,L}\) with a hierarchical dependency such that
    \(\fpos{K}{h,l}(\{\fpos{K}{h,l-1}\}_{h=1}^{H^{l-1}})\). Each of those functions follows the generative process:
    \begin{equation}
\begin{rcases} 
\fnull{h,l} \sim \GP (\mu(\cdot), K(\cdot, \cdot)); &  \W{h,l}\sim p^l_\lambda(\W{h,l}) \\
\btheta = c^l(\Xsamples, \W{h,l}); &\fpos{K}{h,l} \mid \btheta,\fpos{0}{h,l} = \Gcompc{h,l}(\fpos{0}{h,l})
\end{rcases}
\end{equation}
where \(\fnull{0} = \Xsamples\), \(c\) is a transformation that generates coefficients for the normalizing flow $\Gcompcinpdep{h,l}{\Xsamples}$, and \(h=1,\dots,H^l\) is the depth of each layer. By this construction, the joint distribution is given by:
\begin{equation}
\begin{split}
    p(\{\fK{h,l}\}_{h=1,l=1}^{H^l,L}) &= \myprod{l=1}{L} \myprod{h=1}{H^l}\Ngauss{\fnull{h,l}\mid \mu\left(\{\fK{h,l-1}\}_{h=1}^{H^{l-1}}\right),K\left(\{\fK{h,l-1}\}_{h=1}^{H^{l-1}},\{\fK{h,l-1}\}_{h=1}^{H^{l-1}}\right)} \times\\
    &\Jacobiancinputdep{\fpos{k}{h,l}}{h,l}{\Xsamples}
\end{split}
\end{equation}
\end{definition}
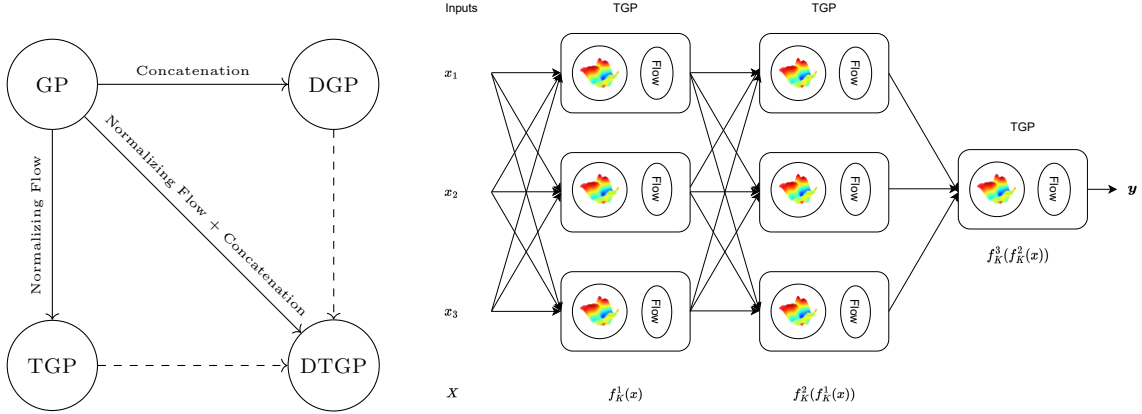
\begin{figure}[t]
	\begin{center}
	\begin{tabular}{cc}
	\begin{tikzpicture}[scale=1.7]
        \tikzstyle{mynode} = [draw, circle, minimum size=1.2cm]
      \node[mynode] (gp) at (0, 0) {\small{\GP}};
      
      \node[mynode] (dgp) at (2.2, 0) {\small{\DGP}};
      \draw[->] (gp) --node[midway, above] {\tiny{Concatenation}}(dgp);
    
      \node[mynode] (tgp) at (0, -2.2) {\small{\TGP}};
      \draw[->] (gp) --node[midway, rotate=90, above] {\tiny{Normalizing Flow}} (tgp);

      \node[mynode] (dtgp) at (2.2, -2.2) {\small{\DTGP}};
      \draw[->] (gp) --node[midway, rotate=-45, above] {\tiny{ Normalizing Flow + Concatenation  }} (dtgp);
      \draw[->, dashed] (tgp) -- (dtgp);
      \draw[->, dashed] (dgp) -- (dtgp);
    \end{tikzpicture}      
 &
	 \includegraphics[width=0.7\textwidth]{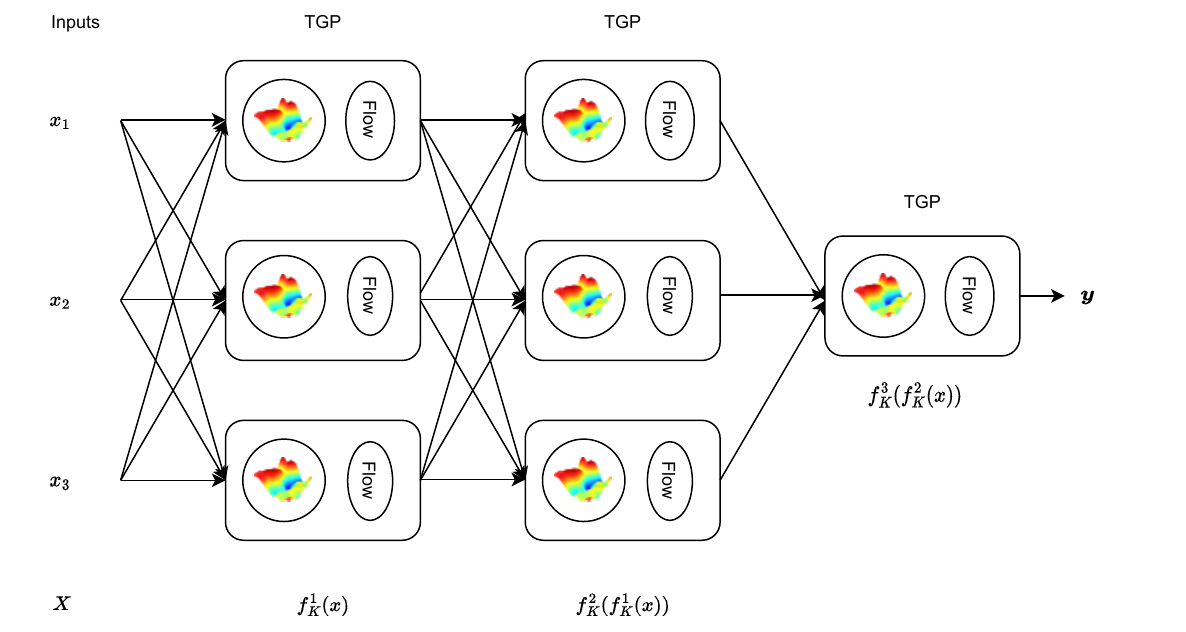}
	\end{tabular}
	\end{center}
\caption{(left): intuitive explanation of how our model was built by combining two generalizations of Gaussian Processes.  (right): Example of \DTGP with 3 layers. In each box, a flow is compounded with the output of a \GP{} (colored circle).} 
    \label{fig:dtgp:structure}

\end{figure}

\fig \ref{fig:dtgp:structure} shows the structure of a \DTGP. Clearly, it can be shown that the proposed architecture is very flexible and contains
as special cases a \GP, a \DGP, and a \TGP, which can all be recovered by selecting an appropriate number of layers and normalizing flows. The usage of normalizing flows between the layers of the \DTGP is expected to allow the model to encode prior expert knowledge about the problem, as it already occurs in \TGP{}s \citep{maronas2021transforming}. From now on, 
we will consider \(H^l = 1\) for all \(l\) to simplify the notation. The subindex \(\btheta^l_h\) will also be omitted as it can be understood from the context. With these assumptions, the joint prior distribution of a \DTGP{} is given by:
\begin{equation}\label{eq:dtgp:prior}
    p(\Ysamples,\deeptransformedset) = \underbrace{\myprod{n=1}{N} p(y_n \mid\Gcompc{L} (\fpos{0,n}{L}))}_{\text{Likelihood}}\underbrace{\myprod{l=1}{L} p(\fnull{l} \mid  \unull{l}) \J{\fK{l} \mid \uK{l}} p(\unull{l}) \J{\uK{l}}}_{\text{ \DTGP Prior}},
\end{equation}
where the diagonal Jacobian appears in each of the layers, representing the transformation of the prior as the novelty from \DGP{}'s prior definition. Since the posterior distribution is intractable, a variational distribution that maintains the exact conditional prior \citep{pmlr-v5-titsias09a} is chosen following \citet{maronas2021transforming}:
\begin{equation}\label{eq:dtgp:variational:distr}
    q(\deeptransformedset)  = \myprod{l=1}{L} p(\fnull{l} \mid \unull{l}) \J{\fK{l} \mid \uK{l}} q(\unull{l})\J{\uK{l}}.
\end{equation}

This choice of variational distribution allows for term cancellations in the \ELBO which implies a gain in computational cost. In virtue of the \emph{Law of the Unconscious Statistician} (\LOTUS), the expression of the \ELBO is the following:
\begin{equation}\label{eq:elbo:dtgp}
        \mathcal L_{\text{\DTGP}} = \underbrace{\mysum{n=1}{N} \Expectation{q(\fnull{L} \mid \fK{L-1})\myprod{l=1}{L-1}q(\fK{l} \mid \fK{l-1})
        }{ \log p \left(y_n \mid  \Gcomp^L (\fnull{L})\right)}}_{\text{ELL}} + \underbrace{\mysum{l = 1}{L} \KL{ q(\unull{l})}{ q(\unull{l})}}_{\text{\KLD}}.
    \end{equation}

\noindent Since \(q(\unull{l})\) and \(p(\unull{l})\) are both Gaussian, the \KLD divergence can be computed analytically. The Expected Log Likelihood (\ELL) term must be approximated since the expectation under the variational distribution is intractable. We approximate this term using Monte Carlo samples from \(\myprod{l=1}{L} q(\fK{l} \mid \fK{l-1}, )\). Thus, we achieve computational tractability using two sources of stochasticity as it was done in \citep{Salimbeni2017}. Firstly, the ELL is a sum across data points, so it can be evaluated using mini-batches. Also, Monte Carlo samples are used to approximate this term. This amounts a computational cost of \(\complexity{(NM^2 + M^3 + NLK)(H^1 + \cdots + H^L)}\) which, compared to the cost of \DGP{}s, adds the cost of computing the normalizing flow \(NLK\). As it will be shown later, the inference algorithm (Alg. \ref{alg:dtgp:train}) is a straightforward extension of the one presented in \citep{Salimbeni2017}.

\subsection{Using Bayesian Priors on Flows}

Input Dependent flows compute the parameters of the normalizing flow using a neural network that outputs a set of parameters for each of the flows \(\btheta^l_h = \text{\NN}(\Xsamples, \W{l})\). A Bayesian treatment can be given to these parameters \(\W{l}\). Firstly, it is assumed that the distribution of the parameters of the flows are independent between the layers, that is:
\begin{equation}\label{eq:dtgp:prior:weights:flows}
    p(\setWs) = \myprod{l=1}{L} p(\W{l})
\end{equation}

Now, following the observations in \citep{maronas2021transforming}, the new joint prior model and the variational posterior factorizes as:
\begin{align}
\begin{split}
    p(\Ysamples,\deeptransformedset, \setWs) & = p(\Ysamples,\deeptransformedset \mid \setWs) p(\setWs),\\
     q(\deeptransformedset, \setWs) & = q(\deeptransformedset \mid \setWs) q(\setWs).
     \end{split}
\end{align}

\noindent Using these expressions, the \ELBO in Equation \eqref{eq:elbo:dtgp} slightly changes. A new \KLD term appears, acting as a regularizer for the distribution of the flow parameters. The complete notation \(\G^L_{\btheta^L(\Xsamples,\W{l})}\) is recovered to remark the dependence of the coefficients obtained by \(\btheta\). Then, \ELBO expression takes the form:
    \begin{align}
            \label{eq:dtgp:elbo:flows:prior}
    \begin{split}
        \mathcal L_{\text{\DTGP}} =  \underbrace{\Expectation{q\left( \deeptransformedset, \setWs \right)}{p(\Ysamples \mid  \Gcompcinpdep{L}{\Xsamples} (\fnull{L}))}}_{\ELL}  & - \underbrace{\mysum{l = 1}{L} \KL{ q(\unull{l})}{ p(\unull{l})}}_{\KLD{}_1} \\ 
        & - \underbrace{\mysum{l = 1}{L} \KL{q(\W{l})}{p(\W{l})}}_{\KLD{}_2}\,.
        \end{split}
    \end{align}

\subsection{Predictions}

In this work, we considered a \DTGP that always uses an identity flow in the last layer \(\Gcompc{L}(\fnull{L}) = \fnull{L}\). First, this allows us to compare directly the modeling advantage that the flows provide in the inner layers compared to identity flows (\ie \DGP). Note that since both inference algorithms rely on the same assumptions, we can attribute the performance difference to additional expressiveness provided by the flows, and not to an improved approximation through a better inference algorithm (\eg using \MCMC \citep{havasi2018inference}).

 The aforementioned simplification has another important advantage: the latent function values at the last layer \(\fnull{L}\) remain Gaussian as a consequence of the linear transformation. This simplifies computing expectations w.r.t. the likelihood. More precisely, it leads to a model that has more flexibility than the \DGP{}s due to the normalizing flows between the layers, but that allows closed-form marginalization of the latent function values \(\fnull{L}\) at the last layer (given \(\fK{L-1}\)), as in \DGP{}s. Importantly, note that the only difference between the proposed inference algorithm for \DTGP{}s and the \DSVI algorithm for \DGP{}s relies on the fact that the samples of each of the hidden layers are passed through a non-linearity. Algorithm \ref{alg:dtgp:train} shows the \DTGP \ELBO evaluation, where the remarked part is the added difference with respect to the \DSVI \DGP{}.\\

Having an input \(\X{}\), this input is propagated through the layers and \(S\) Monte Carlo samples are used to approximate \(q(\fpos{K}{L})\), which has the form:
\begin{equation}\label{eq:predict:dtgp:simple:variational:posterior}
q(\fpos{K}{L}) \approx \frac{1}{S} \mysum{s=1}{S} \mathcal N \left(\fpos{K}{L} \mid \varm{}_{qf}\left(\fpos{K,(s)}{L-1}\right), \varS{}_{qf}\left(\fpos{K,(s)}{L-1},\fpos{K,(s)}{L-1}\right)\right),
\end{equation}
where, if we name \(\alpha(\X{}) = K^L(\Zsamples{L}, \Zsamples{L})^{-1}K^L(\Zsamples{}, \X{})\), 
\begin{align}
    \varm{}_{qf}(\X{}) & = m^L(\X{}) + \alpha(\X{})^T(\varm{L} - m(\Zsamples{L}))\\
    \varS{}_{qf}(\X{i},\X{j}) & = K^L(\X{i},\X{j}) -  \alpha(\X{i})^T\left( K^L(\Zsamples{L},\Zsamples{L}) - \varS{L}\right) \alpha(\X{j}), 
\end{align}
where \(m^L, K^L, \Zsamples{L}, \varm{L}, \varS{L}\) denote the mean function, kernel, inducing locations, variational mean and variational variance of the layer \(L\), respectively.

\noindent The same happens with the predictive distribution for the labels \(y\). We can now assume a Gaussian likelihood to obtain:
\begin{align*}
\begin{split}
    p(y \mid \X{}) & = \int q(\fpos{K}{L} \mid \fpos{K}{L-1}) p(y \mid \fpos{K}{L}) \ d \fpos{K}{L}, \\
    & = \frac{1}{S}\sum_{s=1}^S \mathcal N(\fpos{K}{L} \mid\varm{}_{qf}\left(\fpos{K,(s)}{L-1}\right), \varS{}_{qf}\left(\fpos{K,(s)}{L-1},\fpos{K,(s)}{L-1}\right) + \pmb{\sigma}^2 \pmb{I}),
    \end{split}
\end{align*}
where \(\pmb{\sigma}^2\) is the variance of the likelihood and \(\pmb{I}\) is the identity matrix.\\

\begin{center}
\begin{minipage}{0.66\textwidth}

\begin{algorithm}[H]
\caption{\DTGP Training Step using a mini batch of size \(B\). We highlight the differences with \DSVI{} algorithm.}
\label{alg:dtgp:train}
\DontPrintSemicolon
\SetNoFillComment 
{\color{gray}
  \KwData{Receive a batch \(\mathcal D_B = \{(\X{1}, y_1), \cdots, (\X{B}, y_B)\}\)}

  \(\fK{0} \gets \mathcal \{\X{1},\dots, \X{B}\}\)
  
    \tcc{{\color{gray} Iterate through layers}}
  \For{\(l = 1,\dots, L\)}{
    \tcc{{\color{gray} Compute marginal predictive distribution}}
    \(\pmb{\mu}_0^l, \pmb{\Sigma}_0^l,  \gets q(\fnull{l} \mid \fK{l-1})\)
    
    \tcc{{\color{gray} Obtain \(S\) samples from the marginal predictive}}
    \(\fnull{l} \gets Sample(\pmb{\mu}_0^l, \pmb{\Sigma}_0^l, S)\)
    
    {\color{black}
    \If{Input Dependent Flow}{
        \(\btheta^l(\Xsamples, \W{l}) \gets Transformation_l(\Xsamples)\)
    }

    \tcc{{\color{black} Warp the samples using the Normalizing Flow}}
    \(\fK{l} = \G^l_{\btheta^l(\Xsamples, \W{l}}(\fnull{l})\)
    }

  }

    \tcc{{\color{gray} Compute ELL}}
    \(\ELL \gets \frac{N}{B} \mysum{b=1}{B} \Expectation{q(\fK{L})}{\log(p(y_b\mid \fK{L})}\)

    \tcc{{\color{gray} Compute \KLD, analytical form}}
    \(\KLD \gets \mysum{l=1}{L} \KL{q(\unull{l})}{p(\unull{l})} \)

    \tcc{{\color{gray} Compute \ELBO}}
    \(\ELBO \gets \ELL - \KLD\)

    \tcc{{\color{gray} Optimize parameters using automatic differentiation}}
    \(\btheta \gets Optimizer(\ELBO, \btheta)\)
}
\end{algorithm}
\end{minipage}
\end{center}

\section{Related Work}
\label{section4:RW}

The sparse approaches to Gaussian Processes have allowed these models to be computationally tractable when the number of training data points becomes large \citep{GPs-Big-Data}. Some works have also studied the usage of Harmonic Features in \GP{}s and their relation to deep models has been studied by \citet{eleftheriadis2023sparse}.  Deep Gaussian Processes \citep{pmlr-v31-damianou13a} have been extensively studied in recent literature as a generalization of the \GP{}s that increase their flexibility.

Also, the usage of Normalizing Flows \citep{flows_survey} to transform probability density functions has become an active research field \citep{bnkestad2023variational}. Some works warp the \GP{} likelihood to increase its flexibility \citep{NIPS2003_wgp,RIOS2019235}. In our case, we are more interested in the models that warp the prior distribution of the \GP{}, leading to an improvement in  the performance of transformed model \citep{wilson2010copula, NIPS2010_8d317bdc, maronas2021transforming,maronas2022efficient}. Efficient Transformed Gaussian Processes (\ETGP{}s) also increase the speed of standard sparse \GP{}s methods in multi-class classification tasks \citep{maronas2022efficient} so they could be also considered as the base model for concatenation, which we will do in future work. 

Our work builds on \citep{maronas2021transforming} and extends it by applying a concatenation of the Transformed Gaussian Process following the fashion of the previously presented \DGP{}s. This same idea has been used in different works that use different generalizations of \GP{}s to define more complex models. In \citep{ortega2023deep}, the concatenated model is the Implicit Processes (\IP{}), and in \citep{IDGP:2020} the way the covariance is computed is changed, leading to a more efficient model than \DGP{}s. Our work differs from the previous ones in the form of extending the \GP{}s by using Normalizing flows and also having the possibility of using input-dependent transformations. Since inference remains intractable in our model, we follow a similar approach to \citep{Salimbeni2017}, taking samples from the posterior that in our case are warped using the normalizing flow.

\section{Experiments}
\label{section5:EXPS}

\begin{figure}
    \centering
    \includegraphics[width=0.49\textwidth]{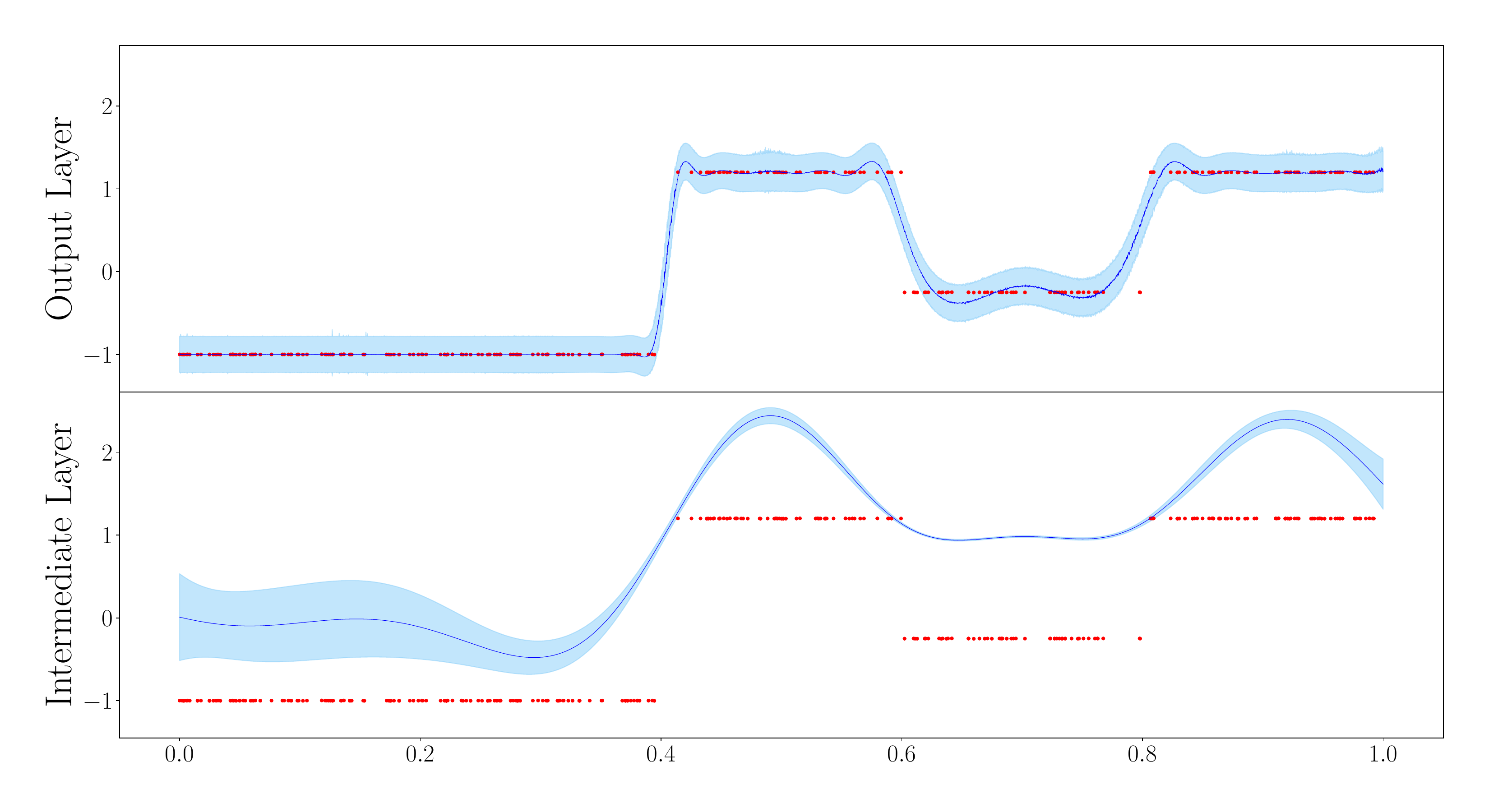}
    \includegraphics[width=0.49\textwidth]{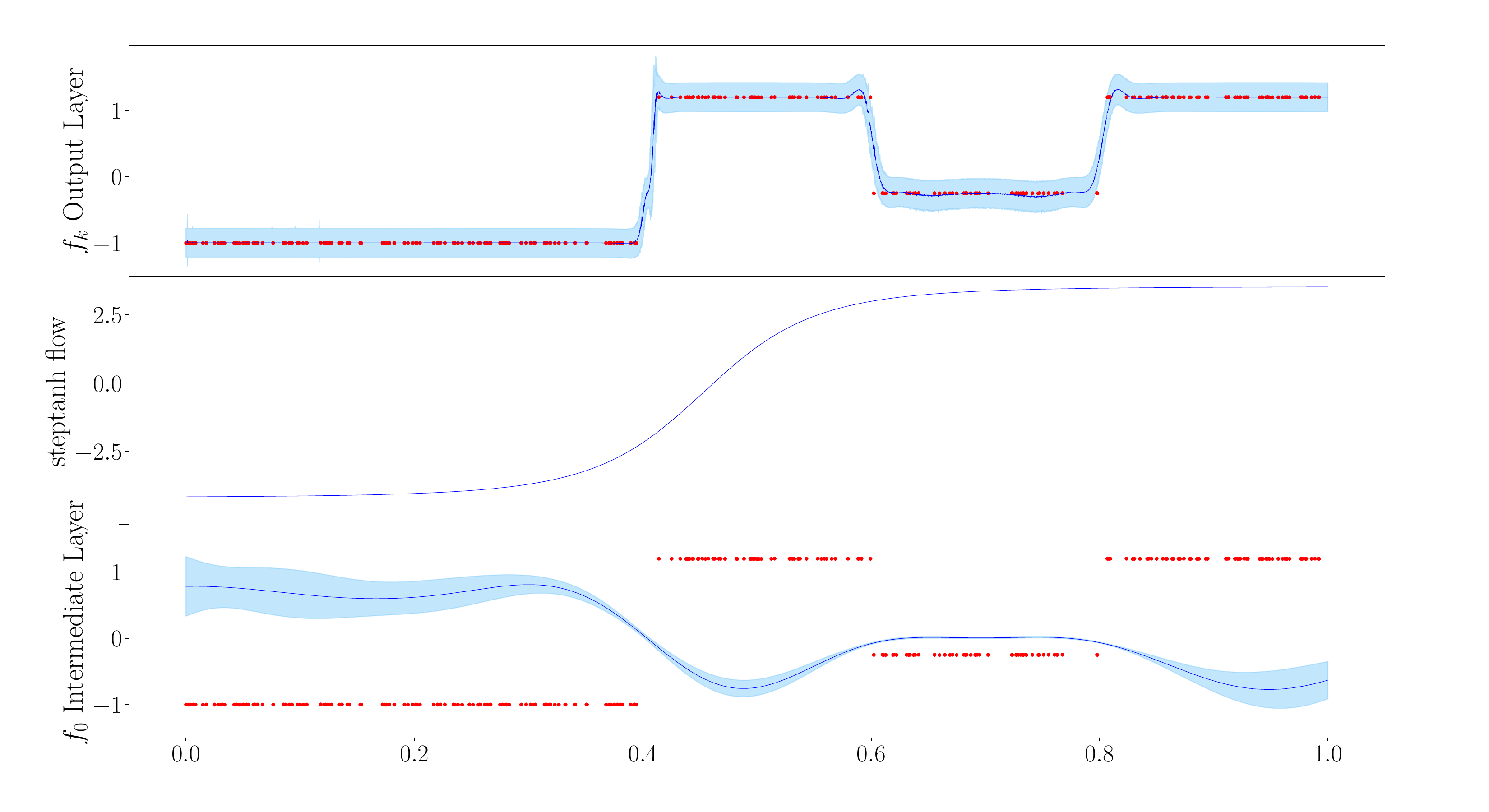}
    \caption{On the left, 2 layer \DGP{}, which has difficulties in modeling the jumps between the steps the data makes. On the right, the proposed \DTGP{} (2 layers) which is capable of fitting the data  better thanks to the nonlinearity in the form of a Steptanh Normalizing flow. }
    \label{fig:comparison:toy}
\end{figure}

\DTGP{}s have been carefully examined in an environment where \DGP{}s do not perform at their best. To this end, \citep{IDGP:2020} presents a toy dataset with a step-wise function where the \DGP{}s have difficulties due to the smoothness that the RBF kernel imposes in the prior. \fig \ref{fig:comparison:toy} shows the visual comparison of \DGP{}s versus \DTGP{}s in this toy dataset. \DTGP{}s use the Steptanh flow, given by a linear combination of the hyperbolic tangent function \citep{maronas2021transforming,maronas2022efficient}. We observe that the proposed model shows not only a better mean prediction of the data but also a good uncertainty estimation where the function value changes. Also, the functional form of the normalizing flow  in \fig \ref{fig:comparison:toy} confirms that using expert information about the problem in the normalizing flows (in this example, a flow that induces a step) can be very helpful for \DTGP{}s.

The last experiments using \DTGP{}s have been conducted using 8 real datasets from the UCI repository. In particular, the datasets used in \cite{Salimbeni2017} have been chosen, exchanging the \emph{Naval} dataset by \emph{Yacht}. The goal of these experiments is to compare the \DTGP{}s with the \DGP{}s, again with the objective of testing if adding normalizing flows in between the layers results in better performance.
As per usual, a \(10\%\) test size has been chosen and each experiment has been performed using \(20\) different random seeds, averaging the results. The initializations of the variational and kernel parameters have followed the ones in \cite{Salimbeni2017}, to replicate their experimental environment.  The number of layers used is 2, 3, 4 and 5 for each of the models, using \(M = 100\) inducing points in each \SVGP. All the models are trained for \(80.000\) iterations using a fixed batch size of \(200\) and a learning rate of \(10^{-2}\). Regarding normalizing flows, it must be remarked that in this work all the layers share the same functional form with different trainable parameters in each layer. Also, we have fixed \emph{noninput dependent arcsinh flow} as the used normalizing flow, since it performed well in some initial experiments. Lastly, \(H_l = 1\) for every \(l=1,\dots,L\). 

\begin{figure}
    \centering
    \includegraphics[scale=0.35]{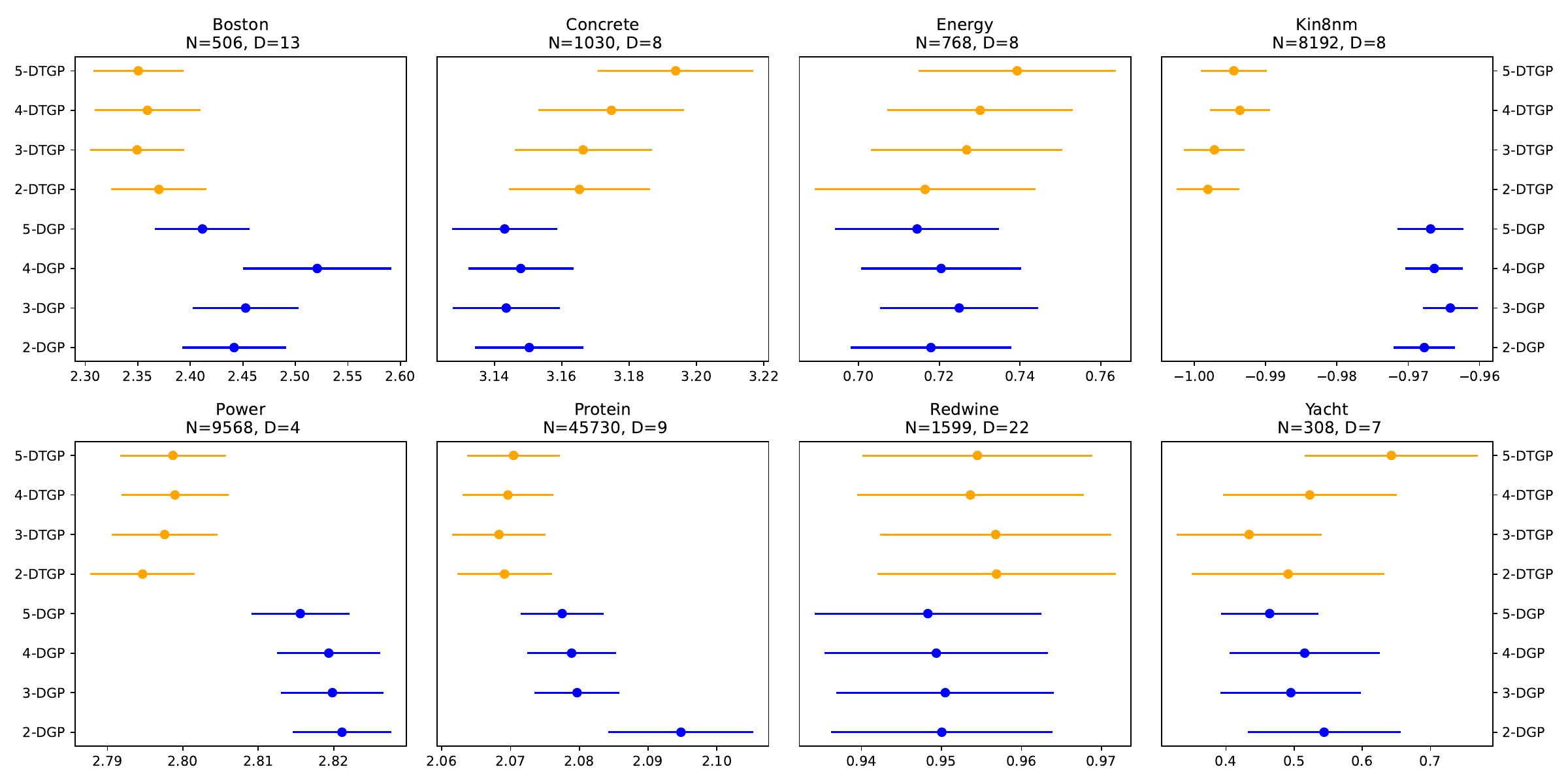}
    \caption{Negative Log Likelihood results (lower values, to the left, are better) comparing \DTGP{}s with noninput dependent Arcsinh Flow with \DGP{}s, both using 2,3,4, and 5 Layers. }
    \label{fig:NLL}
\end{figure}

Using the configuration mentioned above, the results obtained in terms of Negative Log Likelihood (\NLL) for each method are shown in Figure \ref{fig:NLL}. The dot indicates the mean of the \(20\) splits, and the bars indicate the standard deviation divided by \(\sqrt{20}\). It can be observed that  \DTGP{}s achieve better scores in two of the datasets (\emph{Kin8nm} and \emph{Power}), slightly better results in mean in other two (\emph{Boston}, \emph{Protein}), comparable results in three datasets (\emph{Energy}, \emph{Redwine} and \emph{Yacht}) and slightly worse results (in some cases comparable) in \emph{Concrete}. Numeric values are shown in Table \ref{tab:nll:results}.
These results also can give some more insight on \DTGP{}s. While most of the time increasing the number of layers improves the performance in \DGP{}s, in \DTGP{}s sometimes the performance becomes worse. This points out again the difficulty of training the \DTGP{}s since the performance should normally be maintained as the number of layers is increased. Also, the increase in expressiveness may lead to data overfitting, which was already observed in \TGP{}s \citep{maronas2021transforming}. Further work includes Bayesian flows as a way to prevent this overfitting.

\begin{table}[]
    \centering
    \resizebox{\textwidth}{!}{
    \begin{tabular}{ll@{$\pm$}ll@{ $\pm$ }ll@{$\pm$}ll@{$\pm$}ll@{$\pm$}ll@{$\pm$}ll@{$\pm$}ll@{$\pm$}l}
\toprule
   NLL &         \multicolumn{2}{c}{boston} &       \multicolumn{2}{c}{concrete} &         \multicolumn{2}{c}{energy} &          \multicolumn{2}{c}{kin8nm} &          \multicolumn{2}{c}{power} &        \multicolumn{2}{c}{protein} &        \multicolumn{2}{c}{redwine} &          \multicolumn{2}{c}{yacth} \\
\midrule
 2-\DGP{} & 2.442 & 0.049 &  3.150 & 0.016 &  0.718 & 0.02 & -0.968 & 0.004 & 2.821 & 0.007 & 2.095 & 0.011 &  0.950 & 0.014 & 0.544 & 0.112 \\
	    3-\DGP{} &  2.453 & 0.050 & \textbf{3.143} & \textbf{0.016} &  0.725 & 0.020 & -0.964 & 0.004 &  2.820 & 0.007 &  2.080 & 0.006 &  0.950 & 0.014 & 0.495 & 0.103 \\
 4-\DGP{} & 2.521 & 0.071 & 3.148 & 0.016 &   0.720 & 0.020 & -0.966 & 0.004 & 2.819 & 0.007 & 2.079 & 0.006 & 0.949 & 0.014 &  0.516 & 0.110 \\
	    5-\DGP{} & 2.412 & 0.045 & 3.143 & 0.016 &  \textbf{0.714} & \textbf{0.020} & -0.967 & 0.005 & 2.816 & 0.007 & 2.078 & 0.006 & \textbf{0.948} & \textbf{0.0140} & 0.464 & 0.071 \\
 \hline
	    2-\DTGP{} &  2.370 & 0.045 & 3.165 & 0.021 & 0.716 & 0.027 & \textbf{-0.998} & \textbf{0.004} & \textbf{2.795} & \textbf{0.007} & 2.069 & 0.007 & 0.957 & 0.015 & 0.491 & 0.141 \\
	    3-\DTGP{} & \textbf{2.349} & \textbf{0.045} &  3.166 & 0.020 & 0.727 & 0.024 & -0.997 & 0.004 & 2.798 & 0.007 & \textbf{2.068} & \textbf{0.007} & 0.957 & 0.014 & \textbf{0.434} & \textbf{0.107} \\
4-\DTGP{} &  2.359 & 0.05 & 3.175 & 0.022 &  0.730 & 0.023 & -0.994 & 0.004 & 2.799 & 0.007 &  2.07 & 0.007 & 0.954 & 0.014 & 0.523 & 0.127 \\
5-\DTGP{} &  2.350 & 0.043 & 3.194 & 0.023 & 0.739 & 0.024 & -0.994 & 0.005 & 2.799 & 0.007 &  2.070 & 0.007 & 0.955 & 0.014 & 0.643 & 0.127 \\
\bottomrule
\end{tabular}}
    \caption{Negative Log Likelihood exact results.}
    \label{tab:nll:results}
\end{table}

\begin{figure}
    \centering
    \includegraphics[scale=0.375]{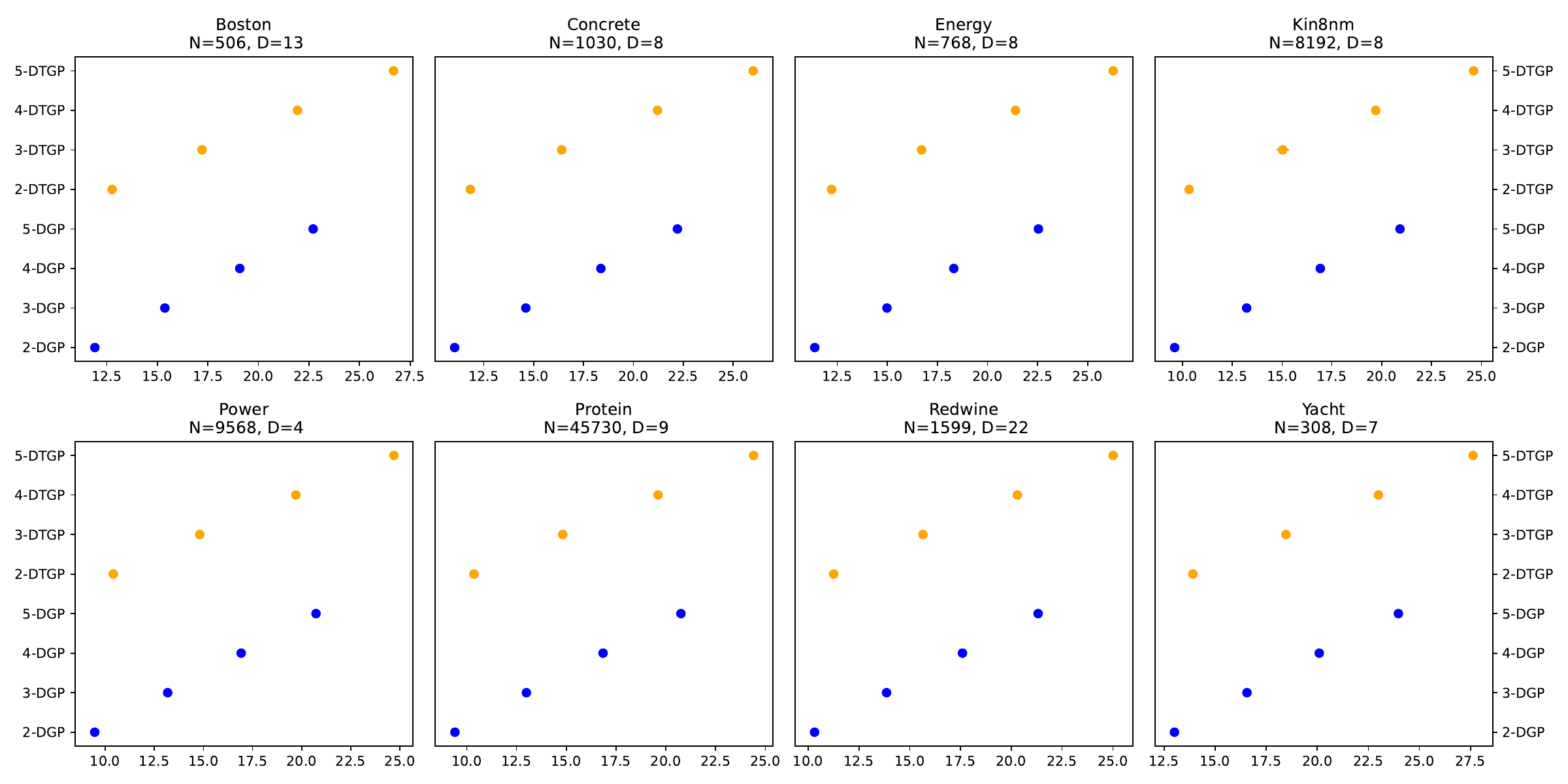}
    \caption{CPU time elapsed by each model in 1000 iterations, averaged in 20 splits.}
    \label{fig:TIME}
\end{figure}

Another aspect that must be carefully examined is the \CPU time used by each method. To measure this, we measure the elapsed time in the first 1000 iterations.
The results are shown in Figure \ref{fig:TIME}. We observe that the training time in seconds becomes more significant as the number of layers grows. 
Specifically, a \DTGP{} with 4 layers needs almost the same computational time as a 5 layer \DGP{}.

\section{Conclusions}\label{section6:CONCLUSIONS}

We have presented Deep Transformed Gaussian Processes (\DTGP{}s), a model based on the concatenation of Transformed Gaussian Processes (\TGP{}s). \DTGP{}s increase the expressiveness of both \TGP{}s and \DGP{}s, by transforming the predictive distribution in each layer. Also, we have derived a further extension of the model where Bayesian priors are placed on the transformation that computes the flow's parameters.

\DTGP{}s inherit the intractability of their base models. Due to this variational inference is used to find an approximation to its true posterior distribution. The derivation of the Evidence Lower Bound for \DTGP{}s has been presented. Furthermore, we have shown how to evaluate this lower bound using Monte Carlo samples. In the performed experiments in the toy data, the improvements that the nonlinearities of the \DTGP{}s offer have been shown using different types of normalizing flows. This improvement is reflected in the real \UCI datasets, where the proposed implementation of the model achieves better or at worst comparable results to the \DGP model.

\section*{Acknowledgments}

The authors gratefully acknowledge the use of the facilities of Centro de Computacion Cientifica (CCC) at Universidad Autónoma de Madrid.  The authors also acknowledge financial support from the Spanish Plan Nacional I+D+i, PID2019-106827GB-I00 and PID2022-139856NB-I00, and from the Autonomous Community of Madrid
(ELLIS Unit Madrid).

\newpage
\appendix

\section{DTGP Derivations}\label{ch:apdx:dtgp:derivations} 
This is a complementary section in which the derivations of the Deep Transformed Gaussian Process model are presented. 

\section*{DTGP Prior}

The DTGP prior in Equation \eqref{eq:dtgp:prior} is obtained as follows:
\begin{align*}
    \setlength{\jot}{10pt}
\begin{WithArrows}
p(\Ysamples,\deeptransformedset) & = p(\Ysamples \mid \deeptransformedset) p( \deeptransformedset) \Arrow {Conditional independence \\ + layer factorization}\\
& = p(\Ysamples{} \mid \fK{L}) \myprod{l=1}{L} p(\fK{l}, \uK{l}) \Arrow {Change of Variable} \\
& = p(\Ysamples \mid \Gcompc{L} (\fnull{L}))\myprod{l=1}{L} p(\fnull{l}, \unull{l}) \J{\fK{l}, \uK{l}} \\
& = \myprod{n=1}{N} p(y_n \mid \Gcompc{L} (\fpos{0,n}{L}))\myprod{l=1}{L} p(\fnull{l} \mid  \unull{l}) p(\unull{l}) \J{\fK{l}, \uK{l}}\\
& = \underbrace{\myprod{n=1}{N} p(y_n \mid\Gcompc{L} (\fpos{0,n}{L}))}_{\text{Likelihood}}\underbrace{\myprod{l=1}{L} p(\fnull{l} \mid  \unull{l}) p(\unull{l}) \J{\fK{l}, \uK{l}}}_{\text{ DTGP Prior}}
\end{WithArrows}
\end{align*}

\section*{Evidence Lower Bound in DTGPs}

Recalling that when using marginal flows \(\J{\fK{}}\J{\uK{}} = \J{\fK{},\uK{}}\), the ELBO in Equation \eqref{eq:elbo:dtgp} is derived as follows:

\[
\setlength{\jot}{20pt}
\begin{WithArrows}
\mathcal L(\Ysamples) & = \Expectation{q\left( \deeptransformedset\right)} { \log \frac{p\left(\Ysamples, \deeptransformedset\right)}{q\left(\deeptransformedset\right)}}\\
& =  \Expectation{q\left( \deeptransformedset\right)} { \log 
\frac{p(\Ysamples \mid \Gcompc{L} (\fnull{L})) \myprod{l=1}{L} \cancel{p(\fnull{l} \mid  \unull{l})} p(\unull{l})\cancel{ \J{\fK{l}, \uK{l}}}}
{\myprod{l=1}{L} \cancel{p(\fnull{l} \mid \unull{l} ) } q(\unull{l}) \cancel{\J{\fK{l},\uK{l}}}  } } \\
& = \Expectation{q\left( \deeptransformedset\right)} { \log p(\Ysamples \mid \Gcompc{L} (\fnull{L}) )} + \mysum{l=1}{L}  \Expectation{q\left(\fK{l}, \uK{l}\right)} {\log \frac{p(\unull{l}) }{ q(\unull{l}) }} \Arrow{LOTUS} \\
& = \Expectation{q\left( \deeptransformedset\right)}{ \log p(\Ysamples \mid \Gcompc{L} (\fnull{L}))} + \mysum{l=1}{L}  \Expectation{q\left(\fnull{l}, \unull{l}\right)}{\log \frac{p(\unull{l}) }{ q(\unull{l}) }} \\
& = \underbrace{\Expectation{q\left( \deeptransformedset\right)}{ \log p(\Ysamples \mid \Gcompc{L} (\fnull{L}) )}}_{\text{Expected Log Likelihood}} - \underbrace{\mysum{l = 1}{L} \KL{ q(\unull{l})}{ p(\fnull{l})}}_{\KLD}
\end{WithArrows}
\]

The Expected Log likelihood term (ELL) requires further development. We operate with the expectation as an integral. Recalling that \( p(\Ysamples \mid \Gc{L}(\fnull{L}))\) factorizes over the data, we can observe that, naming \(\ELL =  \Expectation{Q}{\log p(\Ysamples \mid \Gcompc{L} (\fnull{L}) )}\)
\begin{align*}
    \ELL & = \int q(\deeptransformedset) \log \myprod{n=1}{N}  p(\Ysamples_n \mid  \Gcompc{L} (\fnull{L})) \dd \fK{1,\dots,L} d \uK{1,\dots,L} \\
    & = \int  \log \left( \myprod{n=1}{N} p(\Ysamples_n \mid  \Gcompc{L} (\fnull{L})) \right)\left\{\int \myprod{l=1}{L} p(\fK{l} \mid \uK{l}; \fK{l-1}, \Zsamples{l}) q(\uK{l}) d \uK{1,\dots,L} \right\}\dd \fK{1,\dots,L} \\
    & = \int  \log  \left(\myprod{n=1}{N} p(\Ysamples_n \mid  \Gcompc{L} (\fnull{L})) \right)\left\{\int \myprod{l=1}{L} p(\fnull{l} \mid \unull{l}; \fK{l-1}, \Zsamples{l}) q(\unull{l}) \J{\fK{}} \dd \unull{1,\dots,L} \right\}d \fK{1,\dots,L} \\
    & = \int \log \left(\myprod{n=1}{N} p(\Ysamples_n \mid  \Gcompc{L} (\fnull{L})) \right)\left\{ \myprod{l=1}{L}\int  p(\fnull{l} \mid \unull{l}; \fK{l-1}, \Zsamples{l}) q(\unull{l}) \J{\fK{}} \dd \unull{l} \right\}d \fK{1,\dots,L} \\
    & = \int \log \left(\myprod{n=1}{N} p(\Ysamples_n \mid  \Gcompc{L} (\fnull{L})) \right)\left\{ \myprod{l=1}{L}\int  q(\fnull{l},\unull{l} \mid \fK{l-1}, \Zsamples{l}) \J{\fK{}} d \unull{1,\dots,L} \right\}\dd \fK{1,\dots,L} \\
    & = \int \log \left(\myprod{n=1}{N} p(\Ysamples_n \mid  \Gcompc{L} (\fnull{L}))  \right)\myprod{l=1}{L}  q(\fK{l} \mid \fK{l-1}, \Zsamples{l}) d \fK{1,\dots,L} \\
    & = \mysum{n=1}{N} \int \log p\left(\Ysamples_n \mid  \Gcompc{L} (\fnull{L}) \right) \myprod{l=1}{L}  q(\fK{l} \mid \fK{l-1}, \Zsamples{l}) d \fK{1,\dots,L} \\
    & = \mysum{n=1}{N} \Expectation{\myprod{l=1}{L} q(\fK{l} \mid \fK{l-1}, \Zsamples{l})}{ \log p\left(\Ysamples_n \mid  \Gcompc{L} (\fnull{L})\right)}
\end{align*}

Where, to get from the second to the third equations we have applied the LOTUS rule and, from the third to the fourth equation it is used that each integral only depends on one of the factors.

\section*{Expected Log Likelihood in DTGP using Bayesian Priors}

We can expand the ELL term in Equation \eqref{eq:dtgp:elbo:flows:prior} following the same procedure done in the previous case. In this case, since the flow depends on the sampled weights, we will denote it as \(\Gcompcinpdep{l}{\Xsamples}\) with \(l = 1,\dots, L\). The derivations proceed as follows:
\begin{align*}
    \ELL & = \int q(\deeptransformedset, \setWs) \log \myprod{n=1}{N}  p(\Ysamples_n \mid  \Gcompcinpdep{L}{\Xsamples}  (\fnull{L})) \dd  \fK{1,\dots,L} d \uK{1,\dots,L} d \W{1,\dots,L}\\
    & = \int q(\deeptransformedset \mid \setWs) q(\setWs) \log \myprod{n=1}{N}  p(\Ysamples_n \mid  \Gcompcinpdep{L}{\Xsamples} (\fnull{L})) \dd  \fK{1,\dots,L} d \uK{1,\dots,L} d\W{1,\dots,L}\\
    & = \int q(\setWs)\log \myprod{n=1}{N} p(\Ysamples_n \mid  \Gcompcinpdep{L}{\Xsamples} (\fnull{L}))  \left(\int  q(\deeptransformedset \mid \setWs)   d \uK{1,\dots,L}\right)  \dd \fK{1,\dots,L}  d\W{1,\dots,L}\\
    & = \int q(\setWs)\log \myprod{n=1}{N} p(\Ysamples_n \mid \Gcompcinpdep{L}{\Xsamples} (\fnull{L}))  \left( \myprod{l=1}{L}q(\fK{l} \mid \fK{l-1}, \W{l})\right)  d \fK{1,\dots,L}  d\W{1,\dots,L}\\
    & = \int \log \left(\myprod{n=1}{N}  p(\Ysamples_n \mid  \Gcompcinpdep{L}{\Xsamples} (\fnull{L}))\right) \ \myprod{l=1}{L} q(\W{l}) q(\fK{l} \mid \fK{l-1}, \W{l}) \dd \fK{1,\dots,L}  d\W{1,\dots,L}\\
    & = \mysum{n=1}{N} \int \log \left( p(\Ysamples_n \mid \Gcompcinpdep{L}{\Xsamples} (\fnull{L}))\right) \ \myprod{l=1}{L} q(\W{l}) q(\fK{l} \mid \fK{l-1}, \W{l}) \dd \fK{1,\dots,L}  d\W{1,\dots,L}\\
    & = \mysum{n=1}{N}\Expectation{\myprod{l=1}{L} q(\W{l}) q(\fK{l} \mid \fK{l-1}, \W{l})}{\log p\left( \Ysamples_n \mid  \Gcompcinpdep{L}{\Xsamples} (\fnull{L})\right)}\\
    & \stackrel{(1)}{\approx} \frac{1}{S} \mysum{s=1}{S}\mysum{n=1}{N} \Expectation{\myprod{l=1}{L}  q(\fK{l} \mid \fK{l-1}, \W{l}_s)} {\log  p\left( \Ysamples_n \mid  \Gk{\btheta(\W{L}_{s},\Xsamples)}^{L}(\fnull{L})\right)}
\end{align*}
where the approximation in \((1)\) is done using \(S\) Monte-Carlo Samples \(\W{}_s \sim q(\W{})\).

\bibliography{main}     
\end{document}